\pdfoutput=1
\documentclass[11pt]{article}

\usepackage[final]{acl}
\usepackage{algorithm}
\usepackage{algorithmic}
\usepackage{times}
\usepackage[normalem]{ulem}
\useunder{\uline}{\ul}{}
\usepackage{latexsym}
\usepackage{amsmath}
\usepackage{amssymb}
\usepackage{enumitem}
\usepackage{multirow}
\usepackage[T1]{fontenc}
\usepackage[utf8]{inputenc}

\usepackage{microtype}

\usepackage{inconsolata}

\usepackage{graphicx}

\title{One-for-All Pruning: A Universal Model for Customized Compression of Large Language Models}

\author{
 \textbf{Rongguang Ye},
 \textbf{Ming Tang}\thanks{Corresponding Author.}
\\
Department of Computer Science and Engineering \\
and the Research Institute of Trustworthy Autonomous Systems \\
Southern University of Science and Technology, Shenzhen, China
\\
\small{
  \href{mailto:yerg2023@mail.sustech.edu.cn}{yerg2023@mail.sustech.edu.cn}, 
  \href{mailto:tangm3@sustech.edu.cn}{tangm3@sustech.edu.cn}
}
}

\begin{document} 
\maketitle
\begin{abstract}
Existing pruning methods for large language models (LLMs) focus on achieving high compression rates while maintaining model performance. Although these methods have demonstrated satisfactory performance in handling a single user’s compression request, their processing time increases linearly with the number of requests, making them inefficient for real-world scenarios with multiple simultaneous requests. To address this limitation, we propose a \textbf{Uni}veral Model for \textbf{Cu}stomized \textbf{Com}pression (UniCuCo) for LLMs, which introduces a StratNet that learns to map arbitrary requests to their optimal pruning strategy. The challenge in training StratNet lies in the high computational cost of evaluating pruning strategies and the non-differentiable nature of the pruning process, which hinders gradient backpropagation for StratNet updates. To overcome these challenges, we leverage a Gaussian process to approximate the evaluation process. Since the gradient of the Gaussian process is computable, we can use it to approximate the gradient of the non-differentiable pruning process, thereby enabling StratNet updates. Experimental results show that UniCuCo is 28 times faster than baselines in processing 64 requests, while maintaining comparable accuracy to baselines.
\end{abstract}

% \subsection{References}

\section{Introduction}
Large language model (LLM) compression \cite{ma2023llm,zhu2024survey,yu2024edge} aims to reduce the size and computational demands of pre-trained models while preserving their performance. Among commonly used techniques, pruning \cite{frantar2023sparsegpt,yin2023outlier} reduces the size and complexity of pre-trained models by removing less critical weights or layers while retaining their core functionality. By employing these techniques, LLMs can be deployed efficiently in resource-constrained environments, such as edge devices \cite{tseng2024quip}.
\begin{figure}[t]
    \centering
    \includegraphics[width=\linewidth]{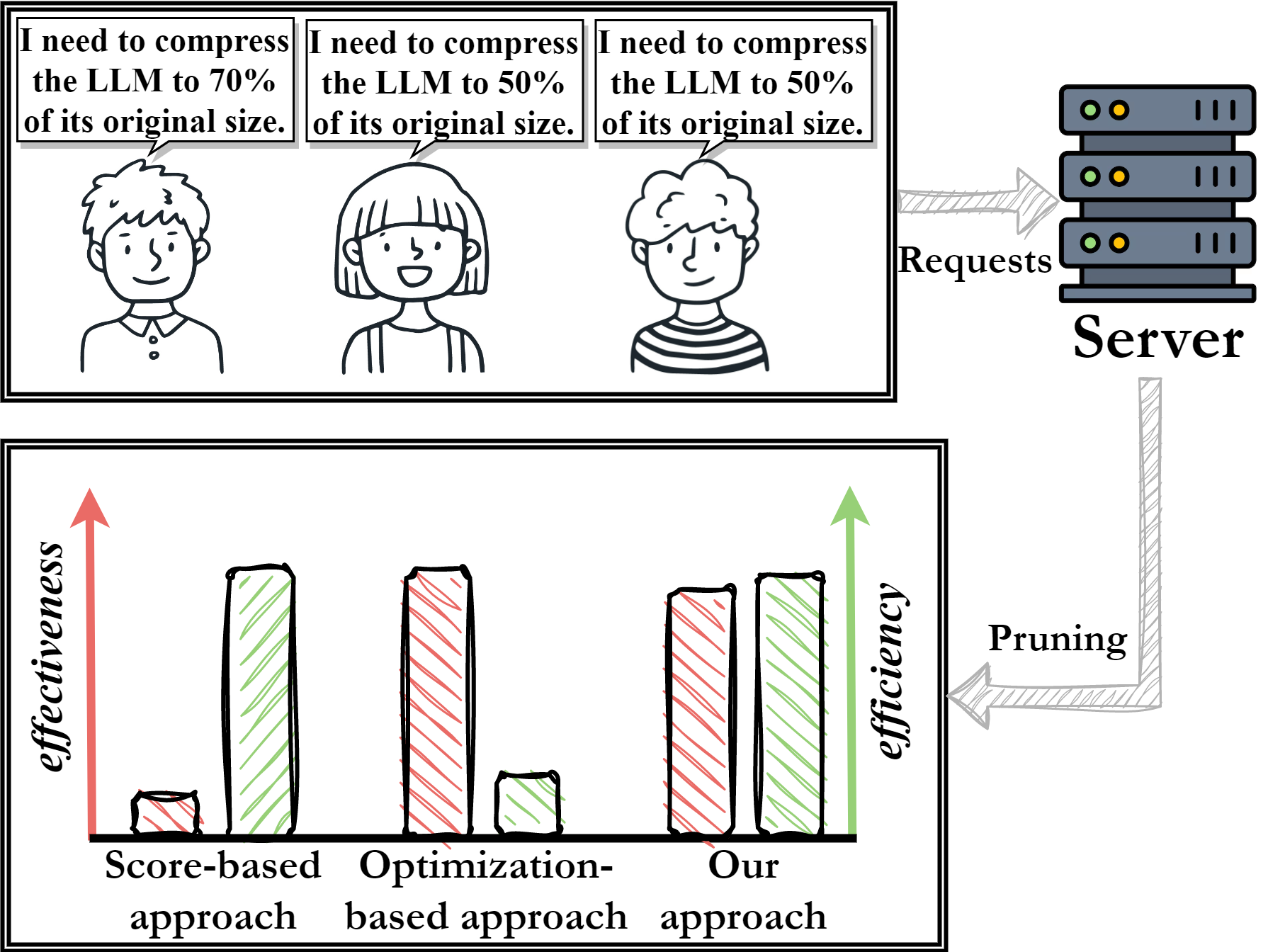}
    \caption{A comparison of various approaches in terms of effectiveness and efficiency when providing pruning strategies for compression requests.}
    \label{moti}
\end{figure}

In practical applications, users have diverse compression requests (defined by their goals of model size reduction while preserving performance) due to the varying capabilities of their devices. Numerous LLM pruning methods have been developed to address specific compression requests \cite{yin2023outlier,kim2024shortened}. These methods can be categorized into optimization-based approaches \cite{sieberling2024evopress} and score-based approaches \cite{men2024shortgpt,frantar2023sparsegpt}. Optimization-based approaches frame LLM pruning as an optimization problem, utilizing heuristic algorithms \cite{yu2010introduction} (e.g., evolutionary algorithms) to preserve the performance of the pruned LLM under a specific compression request. In Fig. \ref{moti}, optimization-based approaches are highly effective in maintaining performance through iterative refinement of pruning strategies during the search process. However, their runtime efficiency in handling multiple compression requests is significantly limited, as each request necessitates an independent heuristic search, leading to substantial time overhead.
In contrast, score-based approaches calculate importance or sensitivity scores for each LLM layer, which are then used to determine the layers to prune to meet compression requests. By reusing the computed scores, the efficiency of handling multiple compression requests is enhanced. However, score-based approaches exhibit limited effectiveness in preserving performance, since the pruning strategies generated from importance scores fail to satisfy the monotonicity property \cite{sieberling2024evopress}.
Motivated by these findings, we pose the following research question: \textbf{How can we \textit{effectively} and \textit{efficiently} handle multiple compression requests?}

To address this problem, we propose Request-Conditional Pruning (UniCuCo) for handling multiple compression requests simultaneously. Specifically, we introduce a StratNet that maps an arbitrary compression request to its corresponding optimal pruning strategy, enabling the handling of diverse requests. It is applicable for StratNet to a wide range of pruning approaches, such as depth pruning and non-uniform pruning. The challenges in training the StratNet are twofold: First, it is difficult to balance the reduction in model size with the preservation of performance when optimizing StratNet to optimally match the user's request. Second, evaluating pruning strategies is computationally expensive. Third, applying pruning strategies (such as binary masks) to the LLM is a non-differentiable operation, which disrupts the backpropagation process and prevents StratNet from being updated using gradient-based methods. To address these challenges, we introduce the weighted Tchebycheff function in the optimization of StratNet, enabling it to effectively derive a pruning strategy that optimally aligns with the given request. Furthermore, we introduce a Gaussian process estimator to evaluate pruning strategies, significantly reducing evaluation time. Since the gradient of the Gaussian process is computable, we leverage it to restore the parts of StratNet that are disrupted in the backpropagation process.
Notably, we propose an alternating update scheme where the Gaussian process and StratNet are updated in an interleaved manner.
The main contributions of this paper are as follows:
\begin{itemize}
\item We introduce the problem of multiple request pruning for LLMs, which requires algorithms to efficiently and effectively generate pruning strategies tailored to diverse requests.
\item We propose UniCuCo, a framework that maps arbitrary compression requests to tailored pruning strategies. UniCuCo includes a Gaussian process estimator, which significantly reduces evaluation time of pruning strategies and subtly solves the non-differentiable issue in UniCuCo. Additionally, we introduce an alternating scheme for updating the Gaussian process and StratNet.
\item Experimental results show that UniCuCo processes 64 compression requests on the Mistral-7B model with a speed at least 28 times faster than optimization-based approaches, while maintaining comparable accuracy. Meanwhile, UniCuCo achieves an average accuracy improvement of 3\% across five benchmark datasets in a non-uniform pruning scenario with 70\% sparsity when compared with score-based approaches on the Mistral-7B model.
\end{itemize}

\section{Related Works}
\subsection{Depth Pruning}
Depth pruning treats each transformer block as a unit and removes entire blocks for pruning. The most common approaches are score-based methods, which compute \textit{block importance} scores and remove those with lower scores based on a compression request. For example, Weight Subcloning \cite{samragh2023weight} is a simple yet effective technique that transfers pre-trained model knowledge to smaller variants by evaluating block importance using the ratio of $\ell_2$ norms between output embeddings with and without residual connections. Shortened LLaMA \cite{kim2024shortened} measures block contribution by removing each block from a pre-trained model and assessing its impact on perplexity. ShortGPT \cite{men2024shortgpt} determines importance through cosine similarity between block inputs and outputs, where lower similarity indicates higher importance. Gromov et al. \cite{gromov2024unreasonable} groups consecutive blocks and evaluates their importance using cosine similarity. However, according to EvoPress \cite{sieberling2024evopress}, score-based approaches in depth pruning are not monotonic. That is, a pruned LLM with a higher cumulative importance score does not necessarily lead to higher effectiveness in preserving performance. To address this limitation, several optimization-based approaches have been proposed for depth pruning. For example, Sheared-LLaMA \cite{xiasheared} introduces a mask learning phase to identify prunable components across blocks.
\begin{figure*}[!t]
    \centering
    \setlength{\abovecaptionskip}{0.05cm}
    \includegraphics[width=\linewidth]{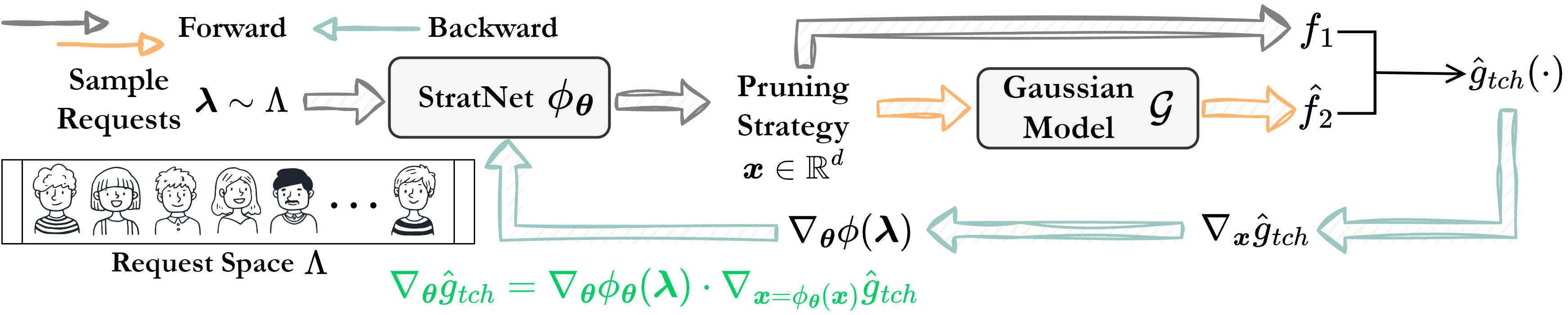}
    \caption{Flowchart of UniCuCo.}
    \label{flowchart}
\end{figure*}
\subsection{Non-Uniform Pruning}
Non-uniform pruning is a more fine-grained pruning scenario, where each transformer block is assigned a sparsity value between 0 and 1, rather than being simply set to 0 or 1. Score-based approaches in non-uniform pruning include Wanda \cite{sunsimple} and SparseGPT \cite{frantar2023sparsegpt}. Wandb evaluates \textit{weight importance} by assessing their impact on the calibration dataset. Specifically, it computes the dot product between the absolute value of the parameter matrix and the $\ell_2$ norm of the input to calculate the weight importance score. SparseGPT computes the Hessian matrix for the weights within each block and generates the corresponding mask matrix based on the compression request. For optimization-based approaches, He et al. \cite{he2018amc} and Ashok et al. \cite{ashok2018n2n} employed reinforcement learning to guide the LLM compression process. However, these approaches are hindered by high computational complexity, leading to significant time overhead when processing a single compression request. To mitigate this issue, the recent OWL method \cite{yin2023outlier} improves compression efficiency by pruning LLMs using layer-wise sparsity ratios proportional to their activation outlier ratios. EvoPress \cite{sieberling2024evopress} formulates compression requests as constraints and employs heuristic search to determine the sparsity of each transformer block. Once the sparsity for each block is determined by EvoPress, SparseGPT \cite{frantar2023sparsegpt} is applied to perform the sparsification of the LLM. Despite these advancements, such methods still incur significant time costs, which are proportional to the number of compression requests. Our work aims to efficiently handle multiple requests while preserving the effectiveness of the pruned LLM.

\section{Methodology}
In this section, we introduce UniCuCo for customized compression of LLMs. Fig. \ref{flowchart} illustrates the flowchart of UniCuCo. First, the core idea is to introduce a StratNet that learns to map arbitrary requests to corresponding pruning strategies (Section \ref{m1}). However, updating the StratNet requires evaluating a large number of pruning strategies on a calibration dataset, which is a time-consuming process.
Thus, we propose the use of a Gaussian process to estimate the evaluation process, thereby reducing the computational overhead of evaluating pruning strategies (Section \ref{m2}).
Finally, we introduce methods for updating StratNet and the Gaussian process (Section \ref{m3}).

\subsection{UniCuCo Framework}\label{m1}
We consider a cloud server that provides pruning strategies $\boldsymbol{x} \in \mathbb{R}^d$ for diverse compression requests, where $d$ is the number of LLM blocks. Each element \( x_i \in [0, 1] \) represents the sparsity ratio of the \( i \)-th block. If \( x_i \) is binary, it corresponds to depth pruning, whereas if \( x_i \) takes a value in the continuous range \( [0, 1] \), it corresponds to non-uniform pruning. Subsequently, we define the compression request and incorporate it into the pruning optimization task. Then, we introduce StratNet and discuss its optimization method.
\subsubsection{Request Formulation}
 Each compression request is associated with \textbf{model size reduction} and \textbf{performance preservation}. To learn the pruning strategy for any given request (i.e., the Pareto front corresponding to all requests), we propose representing compression requests using $\boldsymbol{\lambda} \in \mathbb{R}^2_{+}$, where $\lambda_1 + \lambda_2 = 1$. Each $\lambda_i$ is used to balance the trade-off between model size reduction and performance preservation. The set of all requests over these objectives defines the request space $\Lambda=\{\boldsymbol{\lambda}\in \mathbb{R}^2_{+} \mid \sum_{i=1}^2\lambda_i=1\}$.

Given pruning strategy $\boldsymbol{x}$, we quantify the model size reduction objective as:
\begin{equation}\label{f1}
    \min_{\boldsymbol{x}} f_1(\boldsymbol{x}) = 1 - \frac{\sum_{i=1}^d x_{i}}{d},
\end{equation}
where $\frac{\sum_{i=1}^d x_{i}}{d}$ represents the average sparsity of the pruned LLM. A smaller value of \( f_1(\boldsymbol{x}) \) leads to a smaller model size, as it reflects a higher sparsity.

Following \cite{sieberling2024evopress}, we adopt the KL-divergence between the outputs of the pruned and unpruned LLM to characterize performance preservation:
\begin{equation}\label{f2}
    \min_{\boldsymbol{x}} f_2(\boldsymbol{x}) = {\mathcal{D}}_{\text{KL}}(P_{\mathcal{M}_{\boldsymbol{x}}} \, \| \, P_{\mathcal{M}}),
\end{equation}
where \(P_\mathcal{M}\) denotes the output distribution of the unpruned LLM, while \(P_{\mathcal{M}_{\boldsymbol{x}}}\) is the output distribution of the pruned LLM determined by the pruning strategy \(\boldsymbol{x}\). $f_2$ quantifies the discrepancy in model outputs on the calibration dataset $\mathcal{D}$, with smaller values indicating better performance preservation.

\subsubsection{StratNet}
\begin{figure}[t]
    \centering
    \setlength{\abovecaptionskip}{0.05cm}
    \includegraphics[width=\linewidth]{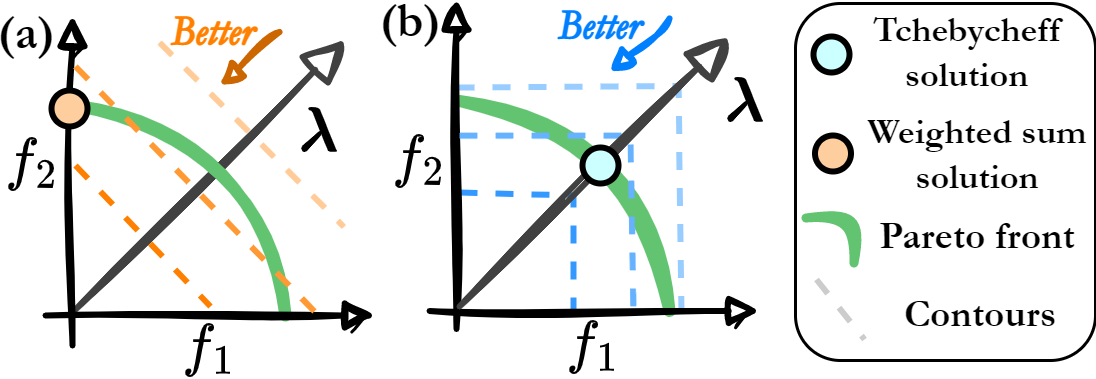}
    \caption{The optimal pruning strategy \( \boldsymbol{x} \) obtained using (a) the weighted sum function and (b) the weighted Tchebycheff function under a concave Pareto front.}
    \label{optimal}
\end{figure}

We propose a StratNet $\phi_{\boldsymbol{\theta}}$, parameterized by $\boldsymbol{\theta}$, that maps the request $\boldsymbol{\lambda}$ to the corresponding pruning strategy $\boldsymbol{x}$, as follows:
\begin{equation}\label{formu}
    \boldsymbol{x}=\phi_{\boldsymbol{\theta}}(\boldsymbol{\lambda}).
\end{equation}
We consider optimizing StratNet with respect to two objectives, $f_1$ and $f_2$. Thus, the optimization of StratNet is formulated as a bi-objective optimization problem: 
$
\min_{\boldsymbol{\theta}} [f_1(\boldsymbol{x}), f_2(\boldsymbol{x})]
$. A straightforward approach to solving this problem is to compute a weighted sum of $f_1$ and $f_2$:
\begin{equation}\label{ws}
    \min_{\boldsymbol{\theta}} g_{ws}(\boldsymbol{x}\mid \boldsymbol{\lambda})=\min_{\boldsymbol{\theta}} \sum_{i=1}^2\lambda_i f_i(\boldsymbol{x}).
\end{equation}
As shown in Fig. \ref{optimal}, \( g_{ws}(\cdot) \) is applicable only to convex Pareto fronts, and not to concave ones. To overcome this limitation, we propose the following weighted Tchebycheff function \cite{miettinen1999nonlinear}:
\begin{equation}\label{tch}
    \!\!\!\! \min_{\boldsymbol{\theta}} g_{tch}(\boldsymbol{x} \mid \boldsymbol{\lambda}) \! = \! \min_{\boldsymbol{\theta}}  \max_{i \in [2]} \left\{ \lambda_i ( f_i(\boldsymbol{x}) - z_i^* ) \right\}, \!\!
\end{equation}
where $z_i^*$ is ideal value for objective $f_i$ (i.e., the lower bound of $f_i$). The function \( g_{tch}(\cdot) \) can be used to identify the optimal pruned strategy in different kinds of Pareto front.

To enable the StratNet to learn pruning strategies for all possible requests, we optimize StratNet over the entire request space $\Lambda$ as follows:
\begin{equation}\label{etch}
\min_{\boldsymbol{\theta}}\mathbb{E}_{\boldsymbol{\lambda}\sim \Lambda} \left[g_{tch}(\boldsymbol{x}=\phi_{\boldsymbol{\theta}}(\boldsymbol{\lambda})\mid \boldsymbol{\lambda})\right].
\end{equation}

For each sampled request $\boldsymbol{\lambda}$ in Eq. \eqref{etch}, \( f_1(\boldsymbol{x}=\phi_{\boldsymbol{\theta}}(\boldsymbol{\lambda})) \) in $g_{tch}$ can be directly computed by Eq. (\ref{f1}). In contrast, $f_2(\boldsymbol{x}=\phi_{\boldsymbol{\theta}}(\boldsymbol{\lambda}))$ cannot be directly computed solely from $\boldsymbol{x}$. Its computation requires generating the compressed model based on $\boldsymbol{x}$ and performing inference on a calibration dataset to derive its value. This introduces two key challenges. 

\noindent (I) \textbf{Significant computational overhead}. For each sampled compression request \(\boldsymbol{\lambda}\), evaluating \(f_2\) with \(g_{tch}(\cdot)\) is time-consuming, as it requires the pruned LLM to perform inferences on the calibration dataset.

\noindent (II) \textbf{The feasibility of updating the StratNet.} The computation of the gradient $\nabla_{\boldsymbol{\theta}}g_{tch}$ is necessary for updating the StratNet. 
However, computing $\nabla_{\boldsymbol{\theta}}g_{tch}$ is challenging because $\nabla_{\boldsymbol{\theta}} f_2$ in $\nabla_{\boldsymbol{\theta}} g_{tch}$ involves the following chain rule:
\begin{equation}\label{chain}
\nabla_{\boldsymbol{\theta}} f_2(\boldsymbol{x}) = \nabla_{\boldsymbol{\theta}} \phi_{\boldsymbol{\theta}}({\boldsymbol{\lambda}}) \cdot \nabla_{\boldsymbol{x}} \mathcal{M}_{\boldsymbol{x}} \cdot \nabla_{\mathcal{M}_{\boldsymbol{x}}} f_2(\boldsymbol{x}),
\end{equation}
where $\nabla_{\boldsymbol{x}} \mathcal{M}_{\boldsymbol{x}}$ cannot be computed, as the mapping from pruning strategy $\boldsymbol{x}$ to a pruned model $\mathcal{M}_{\boldsymbol{x}}$ is a non-differentiable operation. It disrupts the computational chain for Eq. (\ref{chain}).
% The optimization of Problem (\ref{etch}) is challenging due to the infinite number of preference vectors in the entire preference space. To address this, we employ Monte Carlo sampling and gradient descent for optimization:
% \begin{equation}\label{mc}
%     \boldsymbol{\theta} \leftarrow \boldsymbol{\theta} - \eta  \sum_{j=1}^{K}\nabla_{\boldsymbol{\theta}}g_{tch}(\boldsymbol{x}=\phi_{\boldsymbol{\theta}}(\boldsymbol{\lambda}^j)\mid \boldsymbol{\lambda}^j), 
% \end{equation}
% where $K$ is the number of sampled preference vectors and $\eta$ is the learning rate.

\subsection{Gaussian Process for Efficient Estimation}\label{m2}
We introduce a Gaussian process denoted by $\mathcal{G}$, to compute $f_2(\boldsymbol{x})$ when solving problem \eqref{etch}. The key idea is to use $\mathcal{G}$ as an estimator to approximate $f_2(\boldsymbol{x})$ without requiring expensive inference on the calibration dataset for new pruning strategies. This Gaussian process can address challenges (I)--(II). We begin by presenting the Gaussian process and integrating it into our framework.

The Gaussian process $\mathcal{G}$ is defined by a mean function $\mu(\cdot)$ and a covariance function $k(\cdot, \cdot)$:
 \begin{equation}
     \hat{f}_2 \sim \mathcal{G}(\mu(\boldsymbol{x}), k(\boldsymbol{x},\boldsymbol{x})).
 \end{equation}
Initially, we construct a set of $C$ observed samples $\{\boldsymbol{x}^{(c)}, f_2(\boldsymbol{x}^{(c)})\}_{c=1}^{C}$. These samples are used to train the Gaussian process, i.e., to estimate the mean function and covariance function. 

Once $\mathcal{G}$ is trained, it can be used for \textit{inference} on new pruning strategies. When a new pruning strategy $\boldsymbol{x}^{new} = \phi_{\boldsymbol{\theta}}(\boldsymbol{\lambda}^{new})$ is generated by StratNet $\phi_{\boldsymbol{\theta}}$ in problem \eqref{etch}, $\mathcal{G}$ computes the posterior mean $\hat{\mu}(\boldsymbol{x}^{new})$ and variance $\hat{\sigma}^2(\boldsymbol{x}^{new})$ for $\boldsymbol{x}^{new}$. The posterior mean $\hat{\mu}(\boldsymbol{x}^{new})$ serves as an estimate of $f_2(\boldsymbol{x}^{new})$, while the variance $\hat{\sigma}^2(\boldsymbol{x}^{new})$ quantifies the uncertainty of this estimate. To balance exploration (trying uncertain strategies) and exploitation (focusing on good predicted performance), we incorporate uncertainty into the estimation of $f_2$ for $\boldsymbol{x}^{new}$ by employing criteria such as the Lower Confidence Bound (LCB) or Upper Confidence Bound (UCB). Specifically, when applying LCB, the estimate of $f_2$ for $\boldsymbol{x}^{new}$ is given by
\begin{equation}\label{pred}
    \hat{f}_2(\boldsymbol{x}^{new})=\hat{\mu}(\boldsymbol{x}^{new}) + \kappa \hat{\sigma}(\boldsymbol{x}^{new}),
\end{equation}
where $\kappa\geq 0$ is a constant used to balance posterior mean and uncertainty. Using Gaussian process to estimate \( f_2 \) significantly reduces the computation time from tens of seconds on the calibration dataset to just tens of milliseconds, achieving a thousand-fold improvement in computational efficiency and addressing challenge (I).

Meanwhile, the Gaussian process can effectively tackles challenge (II). The existence of Gaussian process ensures that $\nabla_{\boldsymbol{\theta}}f_{2}(\boldsymbol{x})$ can be estimated:
\begin{equation}\label{app}
\nabla_{\boldsymbol{\theta}}\hat{f}_2(\boldsymbol{x})\approx \nabla_{\boldsymbol{\theta}} \phi_{\boldsymbol{\theta}}({\boldsymbol{\lambda}}) \cdot  \nabla_{\boldsymbol{x}}\mathcal{G},
\end{equation}
where $\nabla_{\boldsymbol{x}}\mathcal{G}$ estimates $\nabla_{\boldsymbol{x}}f_{2}(\boldsymbol{x})=\nabla_{\boldsymbol{x}} \mathcal{M}_{\boldsymbol{x}} \cdot \nabla_{\mathcal{M}_{\boldsymbol{x}}}f_2(\boldsymbol{x})$. 
Then, when StratNet is updated, its gradient \( \nabla_{\boldsymbol{\theta}} g_{tch} \) can be estimated by
\begin{equation}
\small
\nabla_{\boldsymbol{\theta}} \hat{g}_{tch}(\boldsymbol{x} \mid \boldsymbol{\lambda}) = 
\begin{cases}
\lambda_1 \nabla_{\boldsymbol{\theta}} f_1(\boldsymbol{x}) & \text{if } \lambda_1 f_1(\boldsymbol{x}) \geq \lambda_2 \hat{f}_2(\boldsymbol{x}), \\
\lambda_2 \nabla_{\boldsymbol{\theta}} \hat{f}_2(\boldsymbol{x}) & \text{if } \lambda_1 f_1(\boldsymbol{x}) < \lambda_2 \hat{f}_2(\boldsymbol{x}).
\end{cases} \nonumber
\end{equation}
In the above equation, when \( \lambda_1 f_1(\boldsymbol{x}) = \lambda_2 \hat{f}_2(\boldsymbol{x}) \), a subgradient is employed, given by \( \lambda_1 \nabla_{\boldsymbol{\theta}} f_1(\boldsymbol{x}) \).

\begin{table*}[t]
\setlength{\tabcolsep}{4pt} % 调整列间距，默认值是 6pt
\resizebox{\textwidth}{!}{
\begin{tabular}{c|c|ccccc|ccccc}
\hline
\multirow{2}{*}{Sparsity} & \multirow{2}{*}{Method} & \multicolumn{5}{c}{Mistral-7B}                & \multicolumn{5}{c}{{Llama-3-8B}}                    \\ \cline{3-12} 
                                   &                                  & Wiki2 (↓) & C4 (↓) & FW (↓) & \textbf{Avg. (↓)} & \textbf{Latency (↓)}      & Wiki2 (↓) & C4 (↓)   & FW (↓)   & \textbf{Avg. (↓)} & \textbf{Latency (↓)}      \\ \hline
0\%                                & Dense                            & 4.82      & 7.72   & 6.41   & 6.32     & --          & 5.54      & 8.80     & 7.62     & 7.32     & --          \\ \hline
\multirow{5}{*}{12.5\%}           
                                   & Cosine (Window)       & 7.19      & 10.18  & 8.39   & 8.59     & \underline{8s}            & 13.21     & 19.56    & 14.27    & 15.68    & \underline{9s}            \\
                             
                                   & EvoPress                         & \textbf{5.74}      & \textbf{9.07}   & \textbf{7.43}   & \textbf{7.41}     & 26.7m         & \underline{7.68}      & \underline{12.33}    & \underline{10.20}    & \underline{10.07}    & 15.4m         \\
                                   & ShortGPT                         & 7.19      & 10.18  & 8.39   & 8.59     & \textbf{\textless{}1s} & 13.21     & 19.56    & 14.27    & 15.68    & \textbf{\textless{}1s} \\
                                         & Weight Subcloning                & 7.19      & 10.18  & 8.39   & 8.59     & \textbf{\textless{}1s} & 13.21     & 19.56    & 14.27    & 15.68    & \textbf{\textless{}1s} \\
                                   & UniCuCo                             & \underline{5.93}      & \underline{9.39}   & \underline{7.67}   & \underline{7.66}     & \textbf{\textless{}1s} & \textbf{7.42}      & \textbf{12.03}    & \textbf{9.80}     & \textbf{9.75}     & \textbf{\textless{}1s} \\ \hline
\multirow{5}{*}{25\%}           
                                   & Cosine (Window)        & 34.94     & 33.7   & 15.08  & 27.91    & \underline{15s}           & 5527.47   & 11588.16 & 2388.11  & 6501.25  & \underline{14s}           \\
                                   
                                   & EvoPress                         & \textbf{10.35}     & \textbf{13.44}  & \textbf{10.63}  & \textbf{11.47}    & 26.2m         & \textbf{14.77}     & \underline{21.30}    & \underline{16.74}    & \underline{17.60}    & 19.6m         \\
                                   & ShortGPT                         & 43.26     & 40.16  & 29.29  & 37.57    & \textbf{\textless{}1s} & 5527.47   & 11588.16 & 2388.11  & 6501.25  & \textbf{\textless{}1s} \\
                                   & Weight Subcloning                & 43.26     & 40.16  & 29.29  & 37.57    & \textbf{\textless{}1s} & 5527.47   & 11588.16 & 2388.11  & 6501.25  & \textbf{\textless{}1s} \\
                                   & UniCuCo                             & \underline{13.73}     & \underline{17.2}   & \underline{13.53}  & \underline{14.82}    & \textbf{\textless{}1s} & \underline{15.05}     & \textbf{20.78}    & \textbf{16.18}    & \textbf{17.34}    & \textbf{\textless{}1s} \\ \hline
\multirow{5}{*}{37.5\%}           
                                   & Cosine (Window)       & 1038.98   & 2362   & 1013.9 & 1471.62  & \underline{15s}           & 64402.73  & 13833.98 & 3908.76  & 27381.82 & \underline{17s}           \\
                                   & EvoPress                         & \textbf{31.91}     & \textbf{30.86}  & \textbf{22.47}  & \textbf{28.41}    & 25.7m         & \textbf{66.21}     & \textbf{80.48}    & \textbf{53.82}    & \textbf{66.84}    & 15.4m         \\
                                   & ShortGPT                         & 2899.74   & 2327.1 & 1023.7 & 2083.50  & \textbf{\textless{}1s} & 64402.73  & 13833.98 & 3908.76  & 27381.82 & \textbf{\textless{}1s} \\
                                   & Weight Subcloning                & 2899.74   & 2327.1 & 1023.7 & 2083.50  & \textbf{\textless{}1s} & 64402.73  & 13833.98 & 3908.76  & 27381.82 & \textbf{\textless{}1s} \\
                                   & UniCuCo                             & \underline{42.48}     & \underline{36.8}   & \underline{25.34}  & \underline{34.87}    & \textbf{\textless{}1s} & \underline{76.89}     & \underline{98.54}    & \underline{55.86}    & \underline{77.10}    & \textbf{\textless{}1s} \\ \hline
\multirow{5}{*}{50\%}           
                                   & Cosine (Window)        & 3410.85   & \underline{1950.6} & 1695.4 & 2352.29  & \underline{10s}           & 2054.46   & 1116.51  & 692.89   & 1287.95  & \underline{12s}           \\
                                   
                                   & EvoPress                         & 4148.65   & 2943.6 & 2937.8 & 3343.32  & 26.3m         & \textbf{496.86}    & \textbf{396.78}   & \textbf{261.37}   & \textbf{385.00}   & 13.0m         \\
                                   & ShortGPT                         & \underline{2423.38}   & 2135.4 & \underline{1104.9} & \underline{1887.89}  & \textbf{\textless{}1s} & 1664.06   & 1739.99  & 1622.69  & 1675.58  & \textbf{\textless{}1s} \\
                                   & Weight Subcloning                & \underline{2423.38}   & 2135.4 & \underline{1104.9} & \underline{1887.89}  & \textbf{\textless{}1s} & 1664.06   & 1739.99  & 1622.69  & 1675.58  & \textbf{\textless{}1s} \\
                                   & UniCuCo                             & \textbf{235.08}    & \textbf{148.85} & \textbf{120.33} & \textbf{168.09}   & \textbf{\textless{}1s} & \underline{983.97}    & \underline{632.06}   & \underline{447.28}   & \underline{687.77}   & \textbf{\textless{}1s} \\ \hline
\multirow{5}{*}{62.5\%}          
                                   & Cosine (Window)        & 8663.29   & 7568.5 & 8644.3 & 8292.01  & \underline{8s}            & \underline{6552.93}   & 2756.67  & \underline{2839.64}  & \textbf{4049.75}  & \underline{9s}            \\
                                   
                                   & EvoPress                         & \underline{3629.51}   & \underline{3039.1} & \underline{2597.9} & \underline{3088.83}  & 28.1m         & \textbf{4711.99}   & \underline{4041.00}  & {4036.07}  & {4263.02}  & 15.2m         \\
                                    & ShortGPT                         & 12539.6   & 10536  & 4755.3 & 9276.92  & \textbf{\textless{}1s} & 56522.08  & 23863.46 & 12350.08 & 30911.87 & \textbf{\textless{}1s} \\
                                    & Weight Subcloning                & 12539.6   & 10536  & 4755.3 & 9276.92  & \textbf{\textless{}1s} & 56522.08  & 23863.46 & 12350.08 & 30911.87 & \textbf{\textless{}1s} \\
                                   & UniCuCo                             & \textbf{1846.28}   & \textbf{1170.4} & \textbf{971.79} & \textbf{1329.49}  & \textbf{\textless{}1s} & 8405.46   & \textbf{2173.26}  & \textbf{1845.45}  & \underline{4141.39}  & \textbf{\textless{}1s} \\ \hline
\end{tabular}}
\setlength{\abovecaptionskip}{0.06cm}
\caption{Depth pruning results of various methods across five sparsity levels, evaluated by perplexity (PPL) and averaged PPL. Latency refers to the time required to handle a single compression request. The best results are highlighted in bold, while the second-best results are underlined.}\label{st1}
\end{table*}

\begin{table*}[t]
\centering
\resizebox{0.94\textwidth}{!}{
\begin{tabular}{cc|cc|ccccc|cc}
 \hline
\multicolumn{1}{c|}{Sparsity}                 & Method   & Wiki2 (↓) & C4 (↓) & ArcC (↑) & ArcE (↑) & HS (↑) & PiQA (↑) & WG (↑) &\textbf{Avg. (↑)} & \textbf{Latency (↓)}      \\ \hline
\multicolumn{1}{c|}{0\%}                      & Dense    & 4.82      & 7.72   & 48.90     & 79.60    & 60.9   & 80.30     & 73.90  & 68.72 & --            \\ \hline
\multicolumn{1}{c|}{\multirow{4}{*}{50\%}} & OWL      & 5.69      & 8.94   & 43.90    & \textbf{76.90}    & 55.4   & \underline{78.50}     & 70.30  & 65.00 & \underline{40m}           \\
\multicolumn{1}{c|}{}                         & EvoPress & \textbf{5.48}      & \textbf{8.69}   & \textbf{44.88}    & 76.85    & \textbf{56.46}  & \textbf{79.16}    & \textbf{71.35}  & \textbf{65.74} & 122m    \\    \multicolumn{1}{c|}{} & Uniform  & 5.68      & \underline{8.93}   & 43.70     & 76.70    & \underline{55.70}   & 78.40     & 71.00   & 65.10 & \textbf{\textless{}1s} \\   
\multicolumn{1}{c|}{}                         & UniCuCo     & \underline{5.65}      & 8.95   & \underline{44.11}    & \underline{76.73}    & 55.66  & 78.40     & \underline{71.27}  & 
\underline{65.23} &\textbf{\textless{}1s} \\ \hline
\multicolumn{1}{c|}{\multirow{4}{*}{60\%}} & OWL      & 7.50       & \underline{11.34}  & \underline{38.50}     & 71.90    & 46.90   & 75.10     & \underline{70.20} & 60.52 & \underline{40m}           \\
\multicolumn{1}{c|}{}                         & EvoPress & \textbf{7.12}      & \textbf{10.91}  & 38.05    & \underline{72.56}    & \textbf{49.91}  & \textbf{76.01}    & 68.98  & \underline{61.10} & 121m    \\  \multicolumn{1}{c|}{} & Uniform  & 7.78      & 11.86  & 38.00       & 72.40    & \underline{49.40}   & 75.00       & 69.30 & 60.82 & \textbf{\textless{}1s} \\
\multicolumn{1}{c|}{}                         & UniCuCo     & \underline{7.44}      & 11.46  & \textbf{39.33}    & \textbf{72.90}    & \textbf{49.91}  & \underline{75.79}    & \textbf{69.93} & \textbf{61.57} & \textbf{\textless{}1s} \\ \hline
\multicolumn{1}{c|}{\multirow{4}{*}{70\%}} 
                        & OWL      & 17.22     & \underline{21.66}  & 27.90     & 62.60    & 38.60   & 67.00       & \underline{63.50}  & 51.92 & \underline{40m}           \\
\multicolumn{1}{c|}{}                         & EvoPress & \textbf{9.73}      & \textbf{14.63}  & \textbf{33.45}    & \textbf{67.13}    & \textbf{43.91}  & \textbf{72.63}    & \textbf{65.27}  & \textbf{56.48} & 120m   \\  \multicolumn{1}{c|}{}  & Uniform  & 23.08     & 30.03  & 27.10     & 60.90    & 36.10   & 65.90     & 59.40   &49.88& \textbf{\textless{}1s}      \\
\multicolumn{1}{c|}{}                         & UniCuCo     & \underline{15.88}     & 22.08  & 29.10     & \underline{64.27}    & \underline{39.05}  & \underline{69.04}    & 62.90  & \underline{52.87}& \textbf{\textless{}1s} \\ \hline
\end{tabular}}
\setlength{\abovecaptionskip}{0.1cm}
\caption{Non-uniform pruning results on the Mistral-7B, evaluated at three sparsity levels, with perplexity for Wiki2 and C4 datasets, and the average zero-shot accuracy (Avg.) across the ArcC, ArcE, HS, PiQA, and WG datasets.}\label{st}
\end{table*}
\subsection{Updating Gaussian Process and StratNet}\label{m3}
The accuracy of the Gaussian process in estimating $f_2$
  is crucial for optimizing StratNet. We present StratNet's update method and a dynamic update approach for the Gaussian process. The Gaussian process is updated once per epoch while StratNet undergoes \( I \) updates per epoch. Their updates are divided into steps (A) and (B). 

\textbf{(A) Initializing Gaussian process and optimizing StratNet.} We randomly initialize a set of $N$ pruning strategies $X^0 = \{\boldsymbol{x}^j\}_{j=1}^N$. Next, we prune the LLM based on $X^0$ and compute the corresponding $f_1$ and $f_2$ values for each pruned LLM. Here, $f_2$ is evaluated on a calibration dataset, while $f_1$ measures the model size reduction. Together, these values form the set $F^0 = \{(f_1(\boldsymbol{x}^j), f_2(\boldsymbol{x}^j))\}_{j=1}^N$. Finally, we train the Gaussian process $\mathcal{G}^0$ on the pairs $\{(\boldsymbol{x}^j, f_2(\boldsymbol{x}^j))\}_{j=1}^N$ by maximizing the marginal likelihood \cite{rasmussen2003gaussian}. 

To optimize StratNet at the first epoch, we apply Monte Carlo sampling to estimate the expectation of requests in Eq. \eqref{mc} and then use gradient descent with $I$ steps for optimization:
\begin{equation}\label{mc}
    \boldsymbol{\theta} \leftarrow \boldsymbol{\theta} - \eta  \sum_{k=1}^{K}\nabla_{\boldsymbol{\theta}}\hat{g}_{tch}(\boldsymbol{x}=\phi_{\boldsymbol{\theta}}(\boldsymbol{\lambda}^k)\mid \boldsymbol{\lambda}^k), 
\end{equation}
where $K$ is the number of sampled requests and $\eta$ is the learning rate.

\textbf{(B) Incremental Gaussian process update and continuous StratNet update.}
During the \(t\)-th epoch, stage (B) selects new samples to expand the training dataset \( \{\boldsymbol{X}^{t-1}, \boldsymbol{F}^{t-1}\} \) in epoch $t-1$ of the Gaussian process, thereby enhancing its prediction accuracy. Then, the Gaussian process updates on the increased training dataset, while StratNet is updated according to Eq. \eqref{mc}.

To increase the training dataset, we first generate a strategy candidate pool \( X^t_p \) based on StratNet. To do this, we sample a set of \( C \) requests \( \{\boldsymbol{\lambda}^c\}_{c=1}^{C} \) from the request space \( \Lambda \). The StratNet then maps these vectors \( \{\boldsymbol{\lambda}^c\}_{c=1}^{C} \) into the candidate pool, i.e., \( \boldsymbol{X}_p^t=\{\boldsymbol{x}^c=\phi_{\boldsymbol{\theta}}(\boldsymbol{\lambda}^c)\}_{c=1}^C \). Next, we compute the $f_1$ of \( \boldsymbol{X}_p^t\) and use the Gaussian process to predict the $f_2$ of \( \boldsymbol{X}_p^t\), resulting in the objective values \( \hat{\boldsymbol{F}}^t_p=(f_1(\boldsymbol{X}_p^t),\hat{f}_2(\boldsymbol{X}_p^t)) \). 

To select a subset from candidate set $\{\boldsymbol{X}_p^t, \hat{\boldsymbol{F}}_p^t\}$ that provides the maximum benefit to Gaussian process training, we use hypervolume (HV) \cite{guerreiro2021hypervolume} to assess the quality of the objective set \( \boldsymbol{F} \). HV is calculated by the area enclosed between each point in \( \boldsymbol{F} \) and a predefined reference point $\boldsymbol{r}$:
\begin{equation}
   \!\!\! \mathcal{H}_{\boldsymbol{r}}(\boldsymbol{F})=\{\boldsymbol{a}\in \mathbb{R}^2 \mid \exists \boldsymbol{f}\in \boldsymbol{F}, \boldsymbol{f} \leq \boldsymbol{a} \leq \boldsymbol{r}\},\!
\end{equation}
where a larger value of $\mathcal{H}_r(\boldsymbol{F})$ reflects a higher quality of the set $F$. The key in improving the Gaussian process lies in measuring the improvement brought by adding the selected subset to the \( (t-1) \)-th epoch training dataset \( \{\boldsymbol{X}^{t-1}, \boldsymbol{F}^{t-1}\} \). Thus, we identify a subset $\{\boldsymbol{X}_s^t, \hat{\boldsymbol{F}}_s^t\}$ from $\{\boldsymbol{X}_p^t, \hat{\boldsymbol{F}}_p^t\}$ with the largest hypervolume improvement ($\mathcal{HI}$) for \( \mathcal{H}_{\boldsymbol{r}}(\boldsymbol{F}^{t-1}) \), as follows:
\begin{equation}\label{hvi}
    \mathcal{HI}(\hat{\boldsymbol{F}}^t_s)=\mathcal{H}_{\boldsymbol{r}}(\boldsymbol{F}^{t-1}\cup\hat{\boldsymbol{F}}_s^t)-\mathcal{H}_{\boldsymbol{r}}(\boldsymbol{F}^{t-1}).
\end{equation}
Based on Eq. \eqref{hvi}, $\boldsymbol{X}_s^t=\arg \max_{\boldsymbol{X}_s^t} \mathcal{HI}(\hat{\boldsymbol{F}}^t_s)$. To add the selected \( \boldsymbol{X}^t_s \) to the \( (t-1) \)-th epoch training dataset \( \{\boldsymbol{X}^{t-1}, \boldsymbol{F}^{t-1}\} \) of the Gaussian process, we need to evaluate \( \boldsymbol{X}^t_s \) on the calibration dataset, as \( f_2 \) of \( \boldsymbol{X}^t_s \) is still estimated by the Gaussian process. To do this, we obtain the pruned LLMs based on \( \boldsymbol{X}^t_s \), and compute their corresponding \( f_1 \) and \( f_2 \) values, forming \( \boldsymbol{F}^t_s \). The training dataset at $t$-epoch is represented as \( \{\boldsymbol{X}^{t}, \boldsymbol{F}^{t}\} = \{\boldsymbol{X}^{t-1} \cup \boldsymbol{X}^t_s, \boldsymbol{F}^{t-1} \cup \boldsymbol{F}^t_s\} \). The Gaussian process is then updated on \( \{\boldsymbol{X}^{t}, \boldsymbol{F}^{t}\} \). 

Afterwards, StratNet performs \(I\) steps of Eq. (\ref{mc}) in \(t\)-epoch.
The pseudocode of UniCuCo is given in Algorithm \ref{alg} of Appendix.

 % CoMC
\section{Experiments}
In this section, we validate the effectiveness and efficiency of our proposed ReCoP against state-of-the-art baselines in both depth pruning and non-uniform pruning scenarios. We further analyze the impact of different scalarization functions (i.e. Eqs. (\ref{ws}), (\ref{tch}) and others) in Appendix \ref{b3}.
\subsection{Experimental Setups}
\noindent \textbf{Baselines.}
For the depth pruning scenario, where the pruning strategy for each layer is represented by binary values (0 and 1), we compare our approach with several baselines. These include the optimization-based method EvoPress \cite{sieberling2024evopress}, as well as score-based methods such as ShortGPT \cite{men2024shortgpt}, Weight Subcloning \cite{samragh2023weight}, and Sliding Window Cosine Similarity (referred toabbreviated as Cosine (Window)) \cite{gromov2024unreasonable}.
\noindent \textbf{Evaluation.}
All competitive methods use Fineweb-Edu (FW) \cite{penedo2024fineweb} as the calibration data. We evaluate \textit{perplexity} on the WikiText-2 (Wiki2) \cite{merity2016pointer} and C4 \cite{raffel2020exploring} datasets to measure the performance of pruned LLMs. Additionally, we assess \textit{accuracy} on zero-shot tasks across a range of datasets, including WinoGrande \cite{sakaguchi2021winogrande}, PiQA \cite{tata2003piqa}, HellaSwag \cite{zellers2019hellaswag}, and both ARC-easy and ARC-challenge \cite{clark2018think}, using the LM Eval Harness \cite{gao2021framework}.

\subsection{Depth Pruning Results}
From the depth pruning results in Table \ref{st1}, two key conclusions can be drawn:
(I) Our proposed UniCuCo outperforms the three score-based methods by delivering pruning strategies in less than one second, similar to the speed of the two fastest score-based methods. Additionally, UniCuCo significantly improves average perplexity, especially as model sparsity increases.
(II) UniCuCo achieves competitive results compared to optimization-based EvoPress, while maintaining significantly shorter latency. Although EvoPress achieves the best perplexity in six out of ten cases across different sparsities and models, it requires approximately 13 to 26 minutes to compute the pruning strategy for each request. In contrast, our proposed UniCuCo takes less than one second in handing each request, while achieving the best results in four out of ten cases. This demonstrates that UniCuCo not only offers fast inference but also remains highly competitive in effectiveness. 
\begin{figure}[t]
    \centering
    \setlength{\abovecaptionskip}{0.05cm}
    \includegraphics[width=\linewidth]{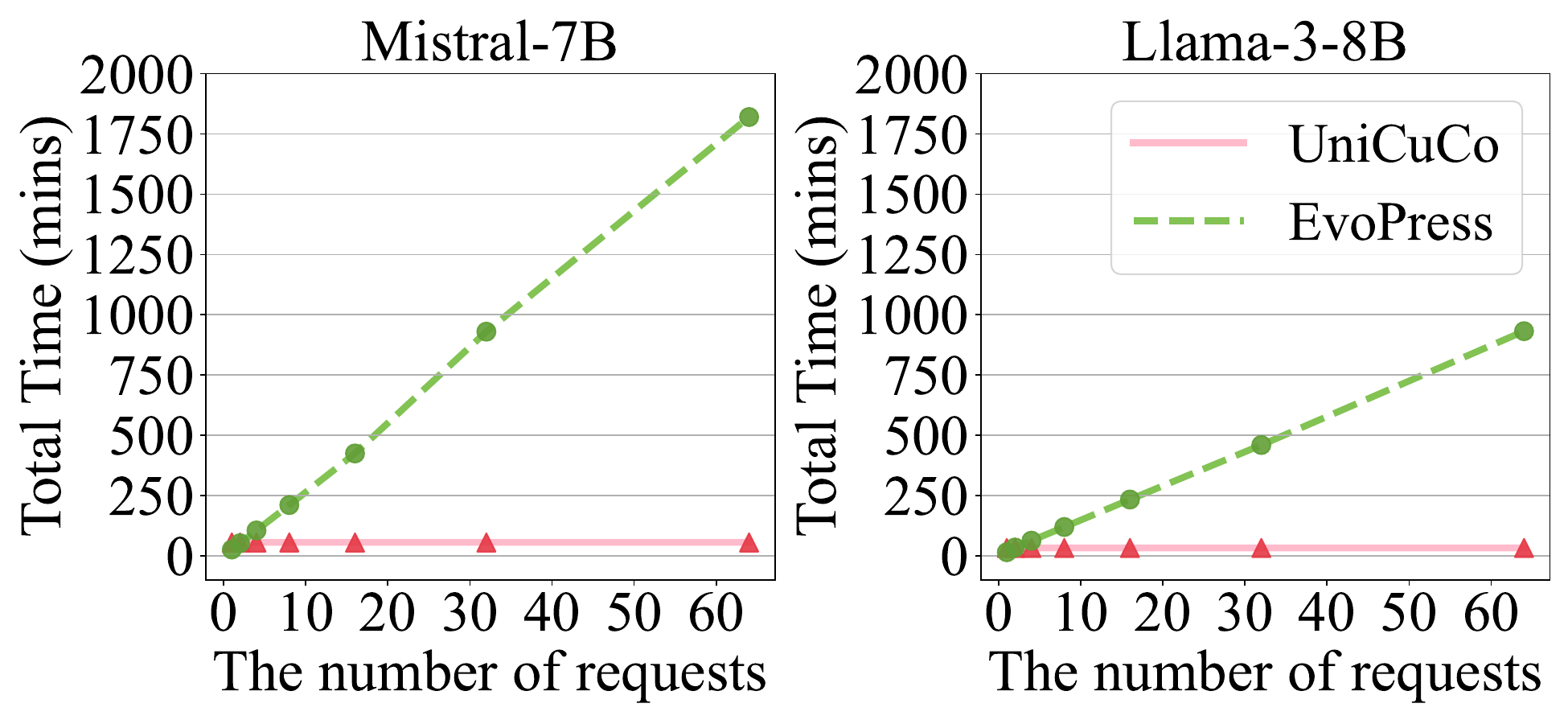}
    \caption{The comparison of total time for generating pruning strategies between UniCuCo and EvoPress as the number of requests increases.}
    \label{numbers}
\end{figure}
\begin{figure}[!t]
    \centering
    \includegraphics[width=0.95\linewidth]{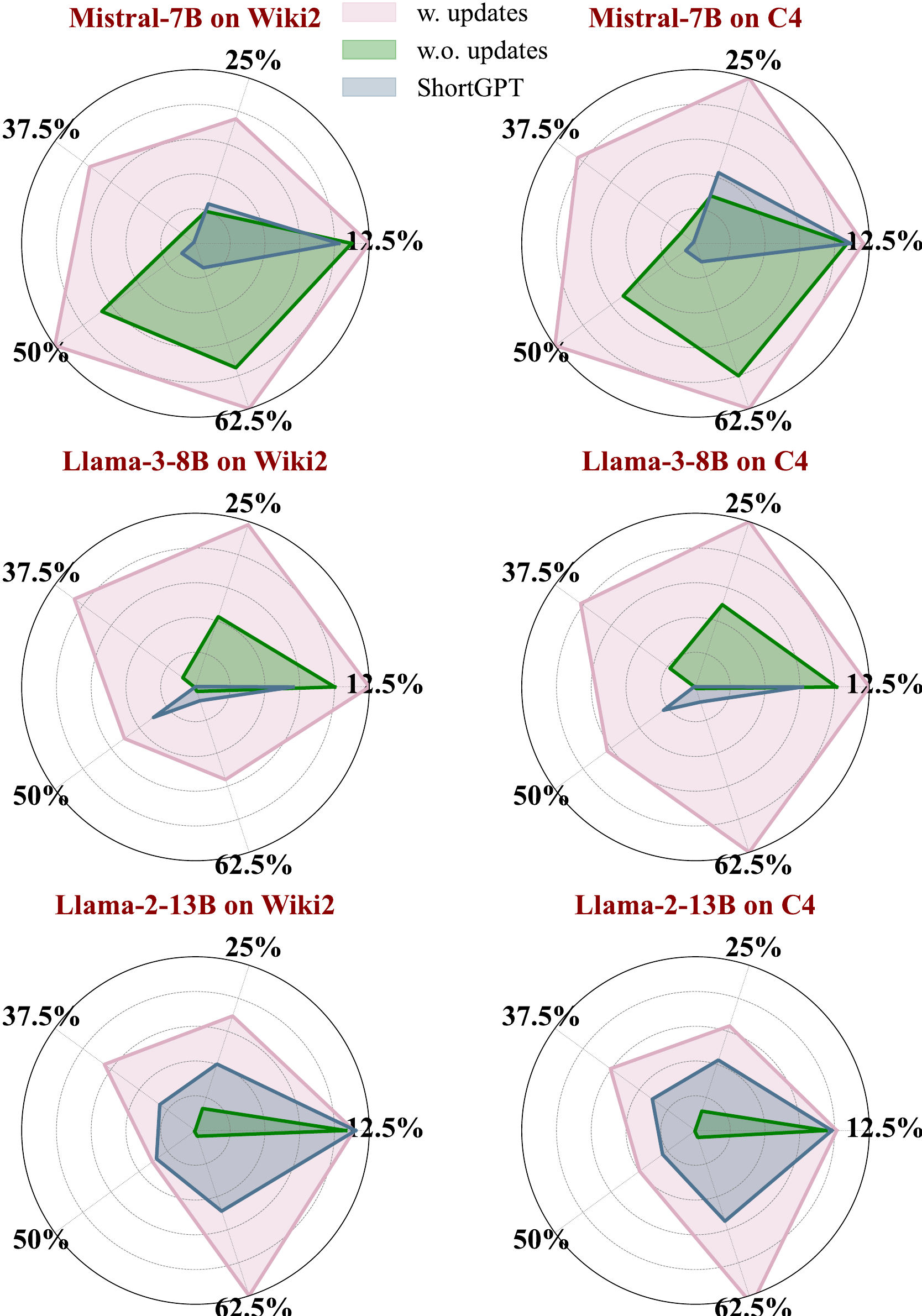}
    \caption{The impact of Gaussian process updates on depth pruning, evaluated using adjusted perplexity (higher values indicate better effectiveness).}
    \label{update}
\end{figure}

We further compare the total time overhead (including both model training time and latency) of UniCuCo and EvoPress in Fig. \ref{numbers}. The results show that as the number of users increases, UniCuCo outperforms EvoPress in total time overhead. Specifically, on Mistral-7B, UniCuCo achieves 56 times more efficiently than EvoPress when scaling to 64 requests. This is because, once the StratNet in UniCuCo is trained, it can generate pruning strategies for any request, whereas EvoPress requires re-searching for each request.

\subsection{Non-Uniform Pruning Results}
In Table \ref{st}, due to the finer granularity of pruning in Non-Uniform Pruning, the distinction between algorithms is less noticeable at low sparsity, while it becomes more significant at 70\% sparsity. In addition, Table \ref{st} shows that our proposed UniCuCo not only achieves lower latency but also outperforms Uniform by 0.1\%, 0.7\%, and 3\% in average accuracy across three sparsities. Furthermore, while EvoPress achieves the best average accuracy in two out of three sparsities, it comes at a prohibitively high time cost for handing single request. Although OWL is more time-efficient than EvoPress, its accuracy is lower by 0.74\%, 0.68\%, and 4.56\% at the three sparsity levels, respectively. In contrast, our proposed UniCuCo achieves a favorable balance, being approximately 2400 times faster than OWL and 7000 times faster than EvoPress per request in terms of efficiency, while outperforming OWL by an average of 0.6\% in accuracy. We provide additional results for non-uniform pruning on Llama-3-8B and Llama-2-13B in Appendix \ref{b2}.
\begin{figure}[t]
    \centering
    \setlength{\abovecaptionskip}{0.05cm}
    \includegraphics[width=\linewidth]{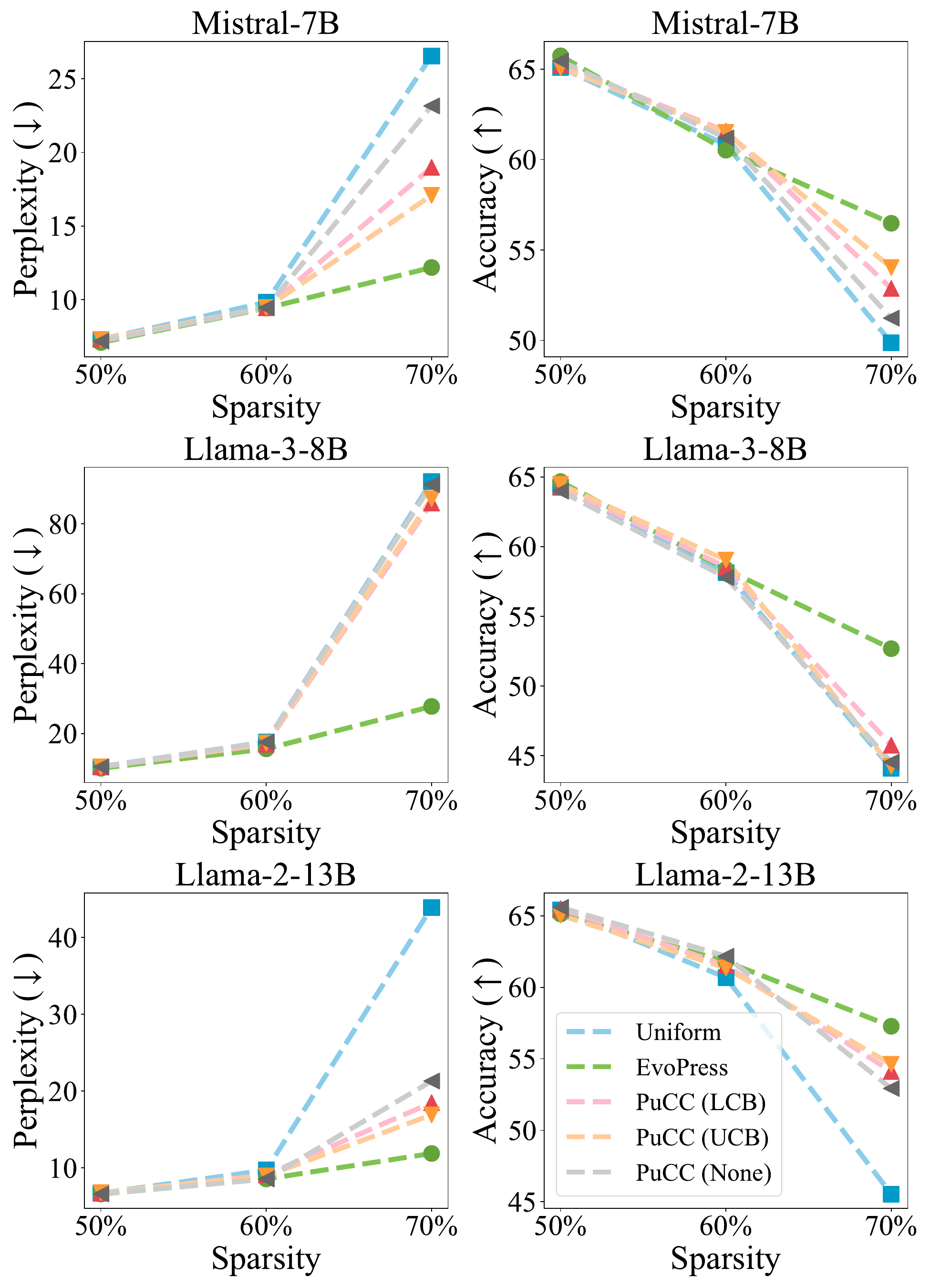}
    \caption{The impact of uncertainty on non-uniform pruning across three large models, evaluated by averaged perplexity and averaged zero-shot accuracy.}
    \label{unc}
\end{figure}
\subsection{Effects of Gaussian Process Updates}
Fig. \ref{update} presents the pruning effectiveness based on normalized perplexity, comparing results with and without Gaussian Process updates. Results show that when the Gaussian Process is not dynamically updated, pruning effectiveness significantly declines compared to when updates are applied. Notably, on Llama-2-13, the absence of Gaussian Process updates leads to effectiveness that is lower than that of the score-based method, ShortGPT. This outcome is intuitive, as StratNet's performance relies on the Gaussian Process's estimation of $f_2$. Insufficient training samples hinder this estimation, thereby degrading the quality of the pruning strategies generated by the StratNet.
\begin{figure}[t]
    \centering
    \setlength{\abovecaptionskip}{0.05cm}
    \includegraphics[width=\linewidth]{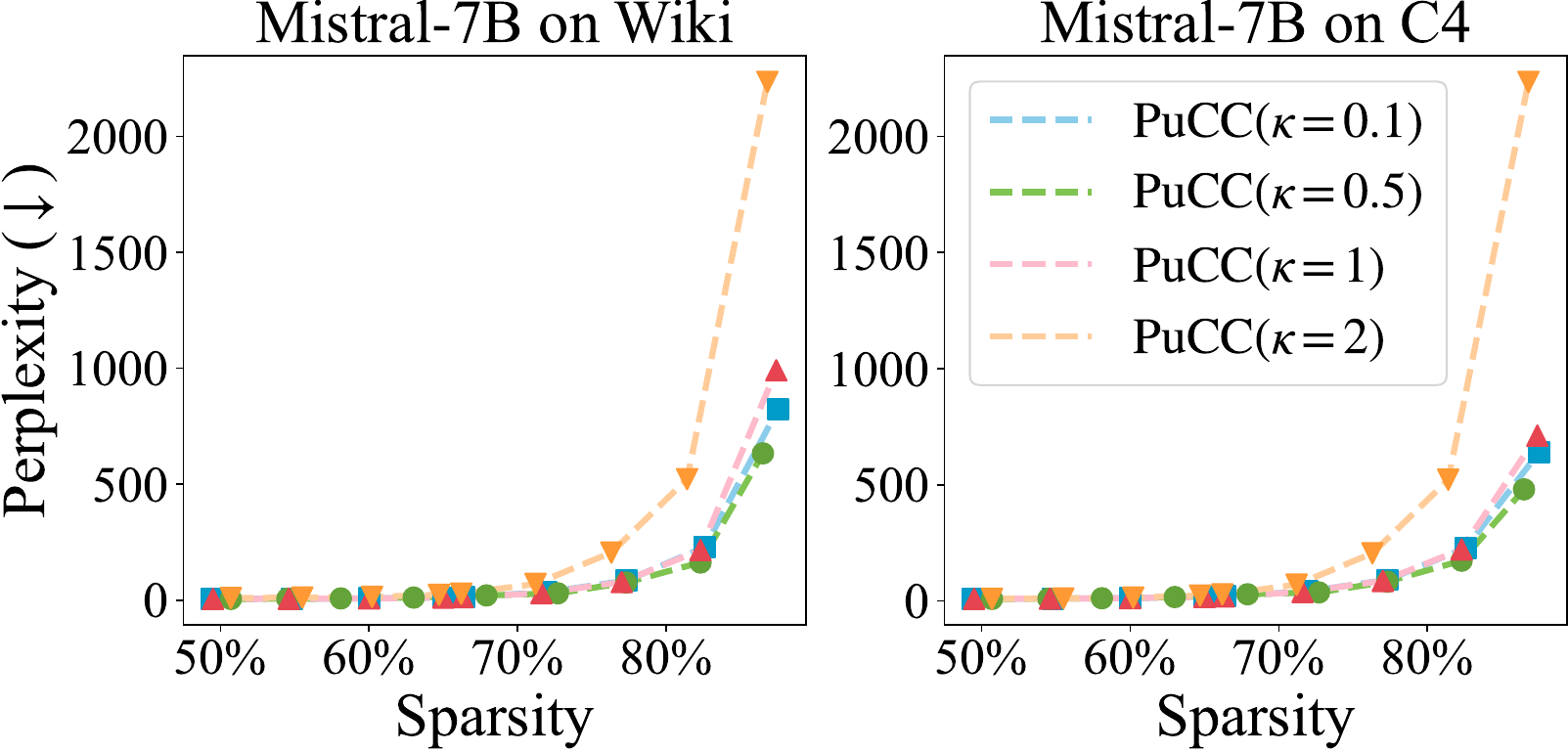}
    \caption{Effect of $\kappa$ on Mistral-7B across two datasets.}
    \label{fig:kappa}
\end{figure}
\subsection{Effects of Uncertainty Estimates}
Recall that in Eq. (\ref{pred}), the prediction of pruning strategies is guided the uncertainty provided by the Gaussian process. Fig. \ref{unc} compares UniCuCo with LCB, UCB, and no uncertainty (None) against Uniform and EvoPress. An interesting observation is that ignoring uncertainty generally leads to worse effectiveness, but it still outperforms Uniform. In contrast, UniCuCo incorporating uncertainty, such as LCB and UCB, delivers superior effectiveness compared to UniCuCo (None). This is intuitively reasonable, as LCB and UCB balance exploiting well-understood regions with exploring areas of higher uncertainty, thereby improving overall effectiveness. Additionally, we show the results on the impact of the uncertainty weight $\kappa$ in Fig. \ref{fig:kappa}. The results indicate that variations in $\kappa$ within the range of 0.1 to 1 have a marginal effect. However, when $\kappa=2$, the pruning perplexity deteriorates as sparsity increases. This is because a larger $\kappa$ causes the Gaussian process to overly emphasize uncertainty, neglecting the predictive mean.

\section{Conclusion}
In this paper, we proposed UniCuCo, an efficient method for handling multiple compression requests while preserving effectiveness. UniCuCo contains a StratNet that learns to map any given request to an optimal compression strategy. To overcome the challenges of high computational cost and gradient incomputability in updating the StratNet, we employ a Gaussian process to approximate updating process, thereby enabling effective learning of the StratNet. Experimental findings indicate that UniCuCo is at least 28 times more efficient than optimization-based methods for processing 64 compression requests. Additionally, it achieves 3\% higher averaged accuracy in a non-uniform pruning scenario with 70\% sparsity when compared with score-based methods.

\section*{Limitations}
Our work currently applies UniCuCo to LLMs with 54 to 80 transformer blocks, ranging from 7B to 13B parameters. The effectiveness of UniCuCo relies on the training of the Gaussian process. As the size of LLMs increases, with hundreds of transformer blocks, the fitting space for Gaussian process training expands, and the pruning strategy dimension in the request space also increases. In this context, the effectiveness of UniCuCo requires further analysis and validation.
Additionally, while our work addresses multiple compression requests for a single LLM, a more complex and realistic scenario involves handling multiple compression requests for multiple LLMs. These aspects will be explored in future research.
% Since December 2023, a "Limitations" section has been required for all papers submitted to ACL Rolling Review (ARR). This section should be placed at the end of the paper, before the references. The "Limitations" section (along with, optionally, a section for ethical considerations) may be up to one page and will not count toward the final page limit. Note that these files may be used by venues that do not rely on ARR so it is recommended to verify the requirement of a "Limitations" section and other criteria with the venue in question.

\section*{Acknowledgments}
This work was supported in part by the National Natural Science Foundation of China under Grant 62202214 and Guangdong Basic and Applied Basic Research Foundation under Grant 2023A1515012819. 

% \clearpage
\bibliography{custom}

\clearpage
\appendix

\section{Experimental Details}\label{appendixa}
\noindent \textbf{Baselines.}
We provide detailed descriptions of the four baseline methods used for comparison in depth pruning as follows:
\begin{itemize}
    \item \textbf{ShortGPT} \cite{men2024shortgpt}: Blocks are scored based on the average cosine similarity between their input and output embeddings, including the residual stream.
    \item \textbf{Weight Subcloning} \cite{samragh2023weight}: Blocks are scored using the ratio $\frac{||\mathcal{M}(\mathcal{E})||}{||\mathcal{M}(\mathcal{E})+\mathcal{E}||}$, where $\mathcal{E}$ is the input embedding and $\mathcal{M}(\mathcal{E})$ is the output of block, excluding the residual stream.
    \item \textbf{Sliding Window Cosine Similarity} \cite{gromov2024unreasonable}: Sets of consecutive blocks are scored based on the cosine similarity between the embeddings before and after the blocks, including the residual stream.
    \item \textbf{EvoPress} \cite{sieberling2024evopress}: Determines whether to discard blocks under given compression constraints using an evolutionary algorithm.
\end{itemize}
For non-uniform pruning, all baselines  and our proposed UniCuCo adopt SparseGPT \cite{frantar2023sparsegpt} as a fast and efficient one-shot layer pruning framework. SparseGPT generates sparsified blocks with varying sparsity levels across layers. The following baselines focus on searching for the optimal sparsity level for each layer:
\begin{itemize}
    \item \textbf{Uniform}: Directly set a uniform sparsity level for all layers and extract the corresponding sparse model generated by SparseGPT.
    \item \textbf{OWL} \cite{yin2023outlier}: OWL uses Layer Outlier Distribution (LOD) metric as a measure of
layer saliency, and computes a sparsity profile that is weighted by LOD.
\end{itemize}
For OWL, we used the same hyperparameter grid as the original
work and took the configuration yielding the best perplexity for each model. Notably, EvoPress can also be applied to non-uniform pruning.

\noindent \textbf{Hyperparameters.} 
Following the setup in \cite{sieberling2024evopress}, the Calibration tokens and Evaluation tokens for the benchmark datasets are set to 524288. The maximum number of tokens that Mistral 7B and Llama-3-8B can process at once is 8192, while Llama-2-13B can process up to 4096 tokens. StratNet employs a multi-layer perceptron (MLP) architecture for training. In the process of increasing the training set in Gaussian model training based on hypervolume contribution, the candidate pool size is set to \( C = 2240 \). In each epoch, \( 10 \) new samples are selected and incorporated into the training set.
 To ensure a fair comparison, we set both EvoPress and UniCuCo to run for 50 epochs (corresponding to $T = 50$ in line 3 of the algorithm \ref{alg}). For the StratNet in UniCuCo, the number of iterations in each epoch is set to 1000 (corresponding to $I = 1000$ in line 6 of the algorithm \ref{alg}) with a learning rate of 1e-3. The Matérn 5/2 kernel is used in the Gaussian process.
\begin{algorithm}[t] 
\caption{UniCuCo algorithm}\label{alg}
\begin{algorithmic}[1] 
\STATE \textbf{Input:} StratNet $\phi_{\boldsymbol{\theta}}$
\STATE \textit{// Initialize the parameters $\boldsymbol{\theta}$ and the training dataset $\{X^0, F^0\}$ for the Gaussian process $\mathcal{G}$}
\FOR{$t=0$ \TO $T$}
    \STATE Training $\mathcal{G}$ on $\{X^t, F^t\}$;
    \STATE Sampled requests $\{\boldsymbol{\lambda}^k\}_{k=1}^K \sim \Lambda$;
    \FOR{$i=0$ \TO $I$}
        \STATE \textit{// StratNet updating}
        \STATE Update $\phi_{\boldsymbol{\theta}}$ by Eq. (\ref{mc});
    \ENDFOR
    \STATE \textit{// Gaussian process updating}
    \STATE Generating a pruning strategy pool \( X_p^t=\{\boldsymbol{x}^c=\phi_{\boldsymbol{\theta}}(\boldsymbol{\lambda}^c)\}_{c=1}^C \);
    \STATE Select a subset $\{X_s^t,F_s^t\}$ based on Eq. (\ref{hvi});
    \vspace{-0.4cm}
    \STATE $\{X^{t}, F^{t}\} \leftarrow \{X^{t-1} \cup X^t_s, F^{t-1} \cup F^t_s\}$;
\ENDFOR
\STATE \textbf{Return:} $\phi_{\boldsymbol{\theta}}$
\end{algorithmic}
\end{algorithm}
\begin{table*}[t]
\centering
\resizebox{\textwidth}{!}{
\begin{tabular}{c|c|ccc|cc}
\hline
Sparsity                                      & Method                     & Wiki2 (↓)       & C4 (↓)          & FW (↓)          & \textbf{Avg. (↓)}       & \textbf{Latency (↓)}              \\ \hline
\multicolumn{1}{c|}{0\%}                      & Dense                      & 4.57            & 6.45            & 5.84            & 5.62            & --                     \\ \hline
\multicolumn{1}{c|}{\multirow{5}{*}{12.5\%}} 
                         & Cosine (Window) & {\ul 5.67}      & 8.09            & 6.97            & {\ul 6.91}      & {\ul 16s}              \\
\multicolumn{1}{c|}{}                         & EvoPress                   & \textbf{5.42}   & \textbf{7.65}   & \textbf{6.62}   & \textbf{6.56}   & 29m                    \\ \multicolumn{1}{c|}{} & ShortGPT                   & 5.88            & 8.36            & 7.19            & 7.14            & \textbf{\textless{}1s} \\
\multicolumn{1}{c|}{}                         & Weight Subcloning          & 5.97            & 8.46            & 7.26            & 7.23            & \textbf{\textless{}1s} \\
\multicolumn{1}{c|}{}              
& UniCuCo                       & 5.94            & {\ul 8.07}      & {\ul 6.94}      & 6.98            & \textbf{\textless{}1s} \\ \hline
\multicolumn{1}{c|}{\multirow{5}{*}{25\%}} 
                         & Cosine (Window)          & {\ul 8.99}      & {\ul 12.57}     & 10.34           & {\ul 10.63}     & {\ul 24s}              \\
\multicolumn{1}{c|}{}                         & EvoPress                   & \textbf{7.20}   & \textbf{9.93}   & \textbf{8.32}   & \textbf{8.48}   & 28.6m   \\           \multicolumn{1}{c|}{} & ShortGPT                   & 17.91           & 19.89           & 15.73           & 17.84           & \textbf{\textless{}1s}  \\  \multicolumn{1}{c|}{}                         & Weight Subcloning          & 17.91           & 19.89           & 15.73           & 17.84           & \textbf{\textless{}1s} \\   
\multicolumn{1}{c|}{}                         & UniCuCo                       & 10.39           & 13.40           & {\ul 10.14}     & 11.31           & \textbf{\textless{}1s} \\ \hline
\multicolumn{1}{c|}{\multirow{5}{*}{37.5\%}} 
                         & Cosine (Window)          & 95.98           & 72.35           & 50.46           & 72.93           & {\ul 24s}              \\
\multicolumn{1}{c|}{}                         & EvoPress                   & \textbf{13.26}  & \textbf{16.69}  & \textbf{13.00}  & \textbf{14.32}  & 26.8m                  \\ \multicolumn{1}{c|}{} & ShortGPT                   & 52.28           & 46.61           & 35.24           & 44.71           & \textbf{\textless{}1s} \\ \multicolumn{1}{c|}{}                         & Weight Subcloning          & 52.28           & 46.61           & 35.24           & 44.71           & \textbf{\textless{}1s} \\
\multicolumn{1}{c|}{}                         & UniCuCo                       & {\ul 20.49}     & {\ul 23.66}     & {\ul 17.74}     & {\ul 20.63}     & \textbf{\textless{}1s} \\ \hline
\multicolumn{1}{c|}{\multirow{5}{*}{50\%}} 
                         & Cosine (Window)          & 1124.49         & 649.33          & 316.09          & 696.64          & {\ul 17s}              \\
\multicolumn{1}{c|}{}                         & EvoPress                   & \textbf{52.06}  & \textbf{35.75}  & \textbf{28.48}  & \textbf{38.76}  & 28.5m                  \\ \multicolumn{1}{c|}{} & ShortGPT                   & 187.99          & 165.59          & 132.47          & 162.02          & \textbf{\textless{}1s} \\ \multicolumn{1}{c|}{}                         & Weight Subcloning          & 187.99          & 165.59          & 132.47          & 162.02          & \textbf{\textless{}1s} \\
\multicolumn{1}{c|}{}                         & UniCuCo                       & {\ul 170.34}    & {\ul 97.43}     & {\ul 76.79}     & {\ul 114.85}    & \textbf{\textless{}1s} \\ \hline
\multicolumn{1}{c|}{\multirow{5}{*}{62.5\%}} 
                         & Cosine (Window)          & 160231.22       & 5219.97         & 4944.90         & 56798.70        & {\ul 13s }                  \\
\multicolumn{1}{c|}{}                         & EvoPress                   & 4437.52         & 3945.83         & 3930.84         & 4104.73         & 23.5m                  \\ \multicolumn{1}{c|}{} & ShortGPT                   & {\ul 1204.75}         & {\ul 864.48}          & {\ul 622.41}          & {\ul 897.23}          & \textbf{\textless{}1s} \\ \multicolumn{1}{c|}{}                         & Weight Subcloning          & {\ul 1204.75}         & {\ul 864.48}          & {\ul 622.41}          & {\ul 897.23}          & \textbf{\textless{}1s} \\
\multicolumn{1}{c|}{}                         & UniCuCo                       & \textbf{588.07} & \textbf{451.34} & \textbf{383.87} & \textbf{474.43} & \textbf{\textless{}1s} \\ \hline
\end{tabular}}
    \caption{Depth pruning results of Llama-2-13B across five sparsity levels, evaluated by validation perplexity (PPL). Avg. represents the averaged perplexity of three datasets.}\label{deap1}
\end{table*}

\begin{table*}[!t]
\centering
\resizebox{\textwidth}{!}{
\begin{tabular}{c|c|cc|ccccc|cc}
\hline
Sparsity                                      & Method   & Wiki2 (↓)      & C4 (↓)         & ArcC (↑)       & ArcE (↑)       & HS (↑)         & PiQA (↑)       & WG (↑)  & \textbf{Avg. (↑)}        & \textbf{Latency (↓)}               \\ \hline
\multicolumn{1}{c|}{0\%}                      & Dense    & 5.54           & 7.10           & 50.40          & 80.10          & 60.20          & 79.70          & 72.60     & 68.60      & --                     \\ \hline
\multicolumn{1}{c|}{\multirow{4}{*}{50\%}} 
                         & OWL      & 8.13           & 13.12          & 43.80          & 75.80          & 54.00          & 75.70          & \textbf{72.20}  & 64.30 & {\ul 30m}              \\
\multicolumn{1}{c|}{}                         & EvoPress & \textbf{7.64}  & \textbf{12.53} & \textbf{43.94} & \textbf{76.18} & \textbf{54.92} & {\ul 76.17}    & {\ul 72.14}   & \textbf{64.67}  & 139m                   \\  \multicolumn{1}{c|}{}  & Uniform  & 8.05           & 13.07          & 43.60          & 75.70          & 54.20          & 76.10          & 71.70    & 64.26       & \textbf{\textless{}1s} \\
\multicolumn{1}{c|}{}                         & UniCuCo     & {\ul 7.97}     & {\ul 12.96}    & {\ul 43.90}    & {\ul 75.84}    & {\ul 54.30}    & \textbf{76.28} & 71.27    & {\ul 64.32}       & \textbf{\textless{}1s} \\ \hline
\multicolumn{1}{c|}{\multirow{4}{*}{60\%}} 
                        & OWL      & \textbf{12.37} & \textbf{18.53} & \textbf{38.00} & \textbf{70.30} & \textbf{47.70} & 72.10          & 68.50    &\textbf{59.22}       & {\ul 30m}              \\
\multicolumn{1}{c|}{}                         & EvoPress & \textbf{12.37} & {\ul 18.92}    & 36.01          & 67.34          & 46.45          & \textbf{72.91} & \textbf{69.14}  & 58.37 & 138m                   \\ \multicolumn{1}{c|}{} 
 & Uniform  & 13.86          & 21.43          & 35.20          & 69.70          & {\ul 45.60}    & {\ul 72.20}    & 68.00    & 58.14       & \textbf{\textless{}1s} \\
\multicolumn{1}{c|}{}                         & UniCuCo     & {\ul 13.29}    & 20.63          & {\ul 36.18}    & {\ul 69.91}    & 46.03          & 72.03          & {\ul 68.51}   & {\ul 58.53}  & \textbf{\textless{}1s} \\ \hline
\multicolumn{1}{c|}{\multirow{4}{*}{70\%}} 
                         & OWL      & {\ul 48.07}    & {\ul 52.32}    & {\ul 27.00}    & {\ul 54.90}    & {\ul 36.60}    & {\ul 65.10}    & {\ul 58.60}   & {\ul 48.44}  & {\ul 30m}              \\
\multicolumn{1}{c|}{}                         & EvoPress & \textbf{24.18} & \textbf{31.38} & \textbf{30.20} & \textbf{61.95} & \textbf{40.03} & \textbf{68.72} & \textbf{62.51} & \textbf{52.68} & 138m                   \\ \multicolumn{1}{c|}{} & Uniform  & 85.84          & 98.35          & 22.70          & 49.90          & 31.40          & 62.10          & 54.40       & 44.10    & \textbf{\textless{}1s} \\
\multicolumn{1}{c|}{}                         & UniCuCo     & 75.12          & 96.82          & 23.46          & 53.66          & 32.36          & 63.11          & 56.20     & 45.76      & \textbf{\textless{}1s} \\ \hline
\end{tabular}}
\caption{Non-uniform pruning results of various methods on the Llama-3-8B model, evaluated at three sparsity levels with validation perplexity (PPL) and zero-shot accuracy. Avg. represents the averaged accuracy of five datasets.}\label{deap2}
\end{table*}

\noindent \textbf{Hardware Details.}
All experiments are conducted on a server running Ubuntu 22.04.5 LTS, equipped with an Intel\textsuperscript{\textregistered} Xeon\textsuperscript{\textregistered} Platinum 8383C CPU @ 2.70GHz and two NVIDIA L40 GPUs (46GB RAM each). All the implementations are performed using the PyTorch framework.

\section{Additional Experiments}
\subsection{Results of UniCuCo on Depth Pruning for LlaMA-2-13B}\label{b1}
Table \ref{deap1} shows that our proposed UniCuCo responds to compression requests in less than one second and outperforms both ShortGPT and Cosine (Window) methods in terms of perplexity, which also respond within one second. Moreover, as sparsity increases, UniCuCo not only achieves competitive results compared to the state-of-the-art method EvoPress but also demonstrates a significant advantage in terms of the time overhead required to handle compression requests over EvoPress.

\begin{table*}[t]
\centering
\resizebox{\textwidth}{!}{
\begin{tabular}{c|c|cc|ccccc|cc}
\hline
Sparsity                                   & Method   & Wiki2 (↓)      & C4 (↓)         & ArcC (↑)       & ArcE (↑)       & HS (↑)         & PiQA (↑)       & WG (↑)  & \textbf{Avg. (↑)}       & \textbf{Latency (↓)}               \\ \hline
\multicolumn{1}{c|}{0\%}                   & Dense    & 4.57           & 6.45           & 49.23          & 77.48          & 79.37          & 80.47          & 72.22    & 71.75       & --                     \\ \hline
\multicolumn{1}{c|}{\multirow{4}{*}{50\%}} 
                      & OWL      & 5.61           & 8.02           & \textbf{45.22} & \textbf{77.23} & {\ul 56.20}    & 77.26          & 72.85    & \textbf{65.75}        & {\ul 114m}             \\
\multicolumn{1}{c|}{}                      & EvoPress & \textbf{5.45}  & \textbf{7.75}  & 42.75          & 76.85          & \textbf{56.49} & 77.69          & 71.90   & 65.14        & 139m                   \\ \multicolumn{1}{c|}{} & Uniform  & 5.54           & 7.90           & 43.17          & {\ul 76.98}    & 55.98          & {\ul 77.86}    & {\ul 73.09}  & 65.42   & \textbf{\textless{}1s} \\
\multicolumn{1}{c|}{}                      & UniCuCo     & {\ul 5.53}     & {\ul 7.89}     & {\ul 43.26}    & 76.94          & 56.02          & \textbf{77.91} & \textbf{73.40}  & {\ul 65.51} & \textbf{\textless{}1s} \\ \hline
\multicolumn{1}{c|}{\multirow{4}{*}{60\%}} 
                     & OWL      & {\ul 7.33}     & {\ul 10.08}    & \textbf{39.85} & {\ul 72.69}    & \textbf{51.08} & {\ul 75.35}    & 69.85       & {\ul 61.76}     & {\ul 98m}              \\
\multicolumn{1}{c|}{}                      & EvoPress & \textbf{7.15}  & \textbf{9.98}  & 39.25          & \textbf{73.11} & {\ul 50.43}    & \textbf{75.52} & \textbf{70.74} & \textbf{61.81} & 138m                   \\ \multicolumn{1}{c|}{} 
 & Uniform  & 8.14           & 11.29          & 38.14          & 72.05          & 48.83          & 74.59          & 69.85         & 60.69  & \textbf{\textless{}1s} \\
\multicolumn{1}{c|}{}                      & UniCuCo     & 7.60           & 10.61          & {\ul 39.33}    & \textbf{73.11} & 49.92          & 74.70          & {\ul 70.56}   & 61.52   & \textbf{\textless{}1s} \\ \hline
\multicolumn{1}{c|}{\multirow{4}{*}{70\%}} 
                      & OWL      & {\ul 14.57}    & {\ul 17.95}    & {\ul 31.31}    & 65.28          & {\ul 41.53}    & {\ul 70.57}    & {\ul 67.09}   & {\ul 55.16}   & {\ul 121m}             \\
\multicolumn{1}{c|}{}                      & EvoPress & \textbf{10.24} & \textbf{13.52} & \textbf{33.87} & \textbf{69.19} & \textbf{44.31} & \textbf{71.76} & \textbf{67.25}  & \textbf{57.28} & 138m                   \\ \multicolumn{1}{c|}{} & Uniform  & 40.33          & 47.5           & 24.06          & 52.19          & 32.17          & 62.13          & 57.14        & 45.54    & \textbf{\textless{}1s} \\
\multicolumn{1}{c|}{}                      & UniCuCo     & 15.85          & 21.13          & 31.23          & {\ul 65.66}    & 39.71          & 69.31          & 64.64        & 54.11    & \textbf{\textless{}1s} \\ \hline
\end{tabular}}
\caption{non-uniform pruning results of various methods on the Llama-2-13B model, evaluated at three sparsity levels with validation perplexity (PPL) and zero-shot accuracy.}\label{deap3}
\end{table*}

\begin{figure*}[t]
    \centering
    \includegraphics[width=0.95\linewidth]{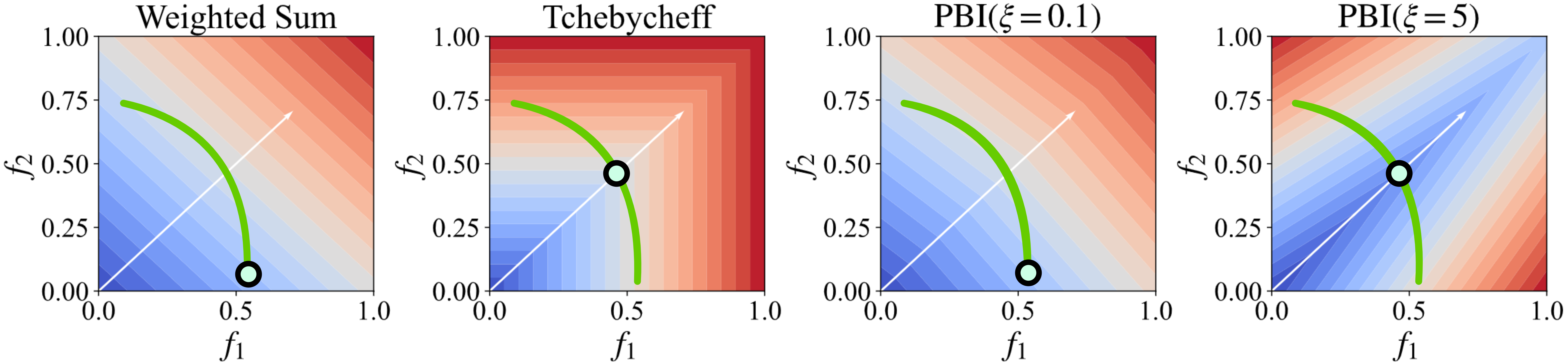}
    \caption{Contour lines of each scalarization function for a two-objective minimization problem. The while arrow represents the request. The green curve and green points represent the Pareto front and the models obtained with different scalarization functions, respectively.}
    \label{fig:agg}
\end{figure*}
\subsection{Results of UniCuCo on Non-Uniform Pruning for Different LLMs}\label{b2}
Tables \ref{deap2} and \ref{deap3} present additional results for non-uniform pruning. We observe that UniCuCo achieves the same time efficiency in handling compression requests as Uniform, while outperforming Uniform in terms of average accuracy. For instance, UniCuCo outperforms Uniform by approximately 1.6\% on Llama-3-8 and by 9\% on Llama-2-13B at a sparsity of 70\%. Furthermore, although UniCuCo slightly lags behind high-demand time algorithms like OWL and EvoPress in terms of effectiveness, it offers an advantage in terms of time efficiency.

\subsection{Impact of Scalarization Function}\label{b3}
As mentioned in Section \ref{m1}, our approach optimizes the StratNet using a weighted Tchebycheff scalarization function (Eq. (\ref{tch})). In this subsection, we compare it with other scalarization functions, such as the weight sum function (Eq. (\ref{ws})) and the Penalty-based Boundary Intersection (PBI) method. The PBI scalarization function is defined as follows:
\begin{equation}
    \min_{\boldsymbol{x}}g_{pbi}(\boldsymbol{x}\mid \boldsymbol{\lambda})=d_1+ \xi d_2,
\end{equation}
where $d_1=\frac{||(\boldsymbol{z}^*-\boldsymbol{f}(\boldsymbol{x}))^{T}\boldsymbol{\lambda}||}{||\boldsymbol{\lambda}||}$ and $d_2=||\boldsymbol{f}-(\boldsymbol{z}^*-d_1\boldsymbol{\lambda})||$. $\xi > 0$ represents a penalty parameter. $d_1$ geometrically represents the distance between $\boldsymbol{f}(\boldsymbol{x})$ and $\boldsymbol{\lambda}$, indicating the degree of alignment with the request. $d_2$ geometrically represents the distance between $\boldsymbol{f}(\boldsymbol{x})$ and the origin when $\boldsymbol{z}^*$ is set to the origin.

\begin{figure}[!h]
    \centering
    \includegraphics[width=0.85\linewidth]{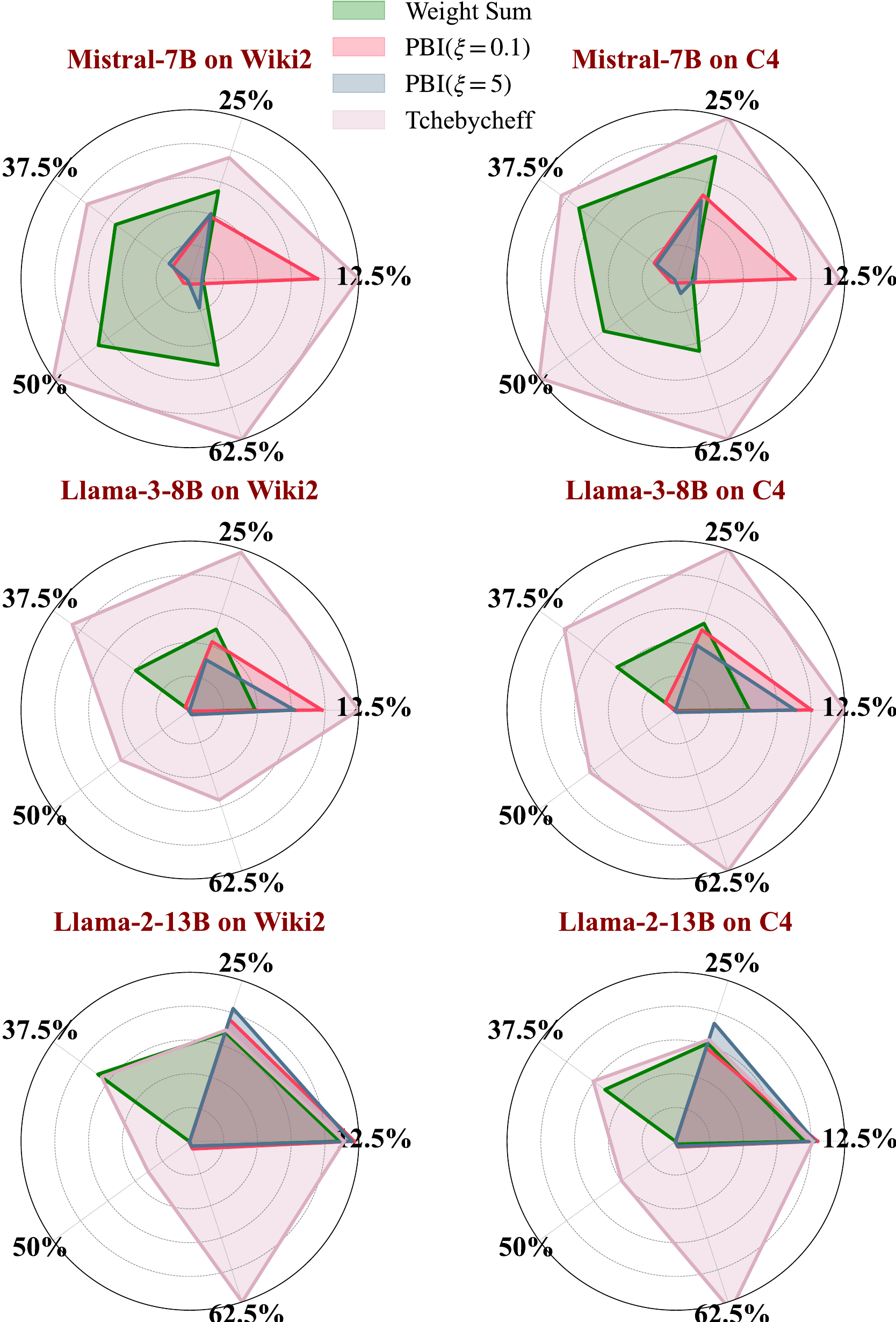}
    \caption{Effects of different scalarization functions on compression performance at five levels of sparsity.}
    \label{fig:agg_result}
\end{figure}

To facilitate the analysis of the properties of the above scalarization functions, we present their contour plots for a given white request $\boldsymbol{\lambda}$ in Fig. \ref{fig:agg}. In these plots, the bluer the color, the smaller the value of the scalarization function, indicating better performance, while the redder the color, the larger the value, indicating poorer performance. Theoretically, the weighted Tchebycheff function can find an optimal solution for any given request, regardless of the shape of the Pareto front. However, the model obtained by weighted sum and PBI ($\xi=0.1$) cannot accurately align request under concave Pareto fronts. For PBI ($\xi=5$), the emphasis on aligning requests (due to the large value of $\xi$) leads to neglecting the scalarization function value, i.e., the distance from the origin. Fig. \ref{fig:agg_result} illustrates the adjusted perplexity (higher values indicate better performance), where the weighted Tchebycheff function achieves the best results, followed by the Weighted Sum and PBI ($\xi=0.1$), with PBI ($\xi=5$) yielding the poorest performance.

\end{document}